\documentclass[sn-mathphys,Numbered]{sn-jnl}

\usepackage[table]{xcolor}
\usepackage{booktabs}
\usepackage{graphicx}%
\usepackage{multirow}%
\usepackage{amsmath,amssymb,amsfonts}%
\usepackage{amsthm}%
\usepackage{mathrsfs}%
\usepackage[title]{appendix}%
\usepackage{textcomp}%
\usepackage{manyfoot}%
\usepackage{booktabs}%
\usepackage{colortbl}
\usepackage{algorithm}%
\usepackage{algorithmicx}%
\usepackage{algpseudocode}%
\usepackage{listings}%
\usepackage{siunitx}
\usepackage{amssymb}
\usepackage{hyperref}
\usepackage{caption}
\usepackage{enumitem}

\theoremstyle{thmstyleone}%
\theoremstyle{thmstyletwo}%

\theoremstyle{thmstylethree}%

\usepackage{color}
\definecolor{forestgreen}{rgb}{0.33,0.61,0.34}
\definecolor{myblue}{rgb}{0, 0.4, 0.8}

\begin{document}

\title{Swarm navigation of cyborg-insects in unknown obstructed soft terrain}


\author[1,2]{\fnm{Yang} \sur{Bai}}
\equalcont{These authors contributed equally to this work.}

\author[3]{\fnm{Phuoc Thanh} \sur{Tran Ngoc}}
\equalcont{These authors contributed equally to this work.}

\author[3]{\fnm{Huu Duoc} \sur{Nguyen}}
\equalcont{These authors contributed equally to this work.}

\author[3]{\fnm{Duc Long} \sur{Le}}

\author[3]{\fnm{Quang Huy} \sur{Ha}}

\author[3]{\fnm{Kazuki} \sur{Kai}}

\author[3]{\fnm{Yu Xiang} \sur{See To}}

\author[1]{\fnm{Yaosheng} \sur{Deng}}

\author[1]{\fnm{Jie} \sur{Song}}

\author*[1]{\fnm{Naoki} \sur{Wakamiya}}
\email{wakamiya@ist.osaka-u.ac.jp}

\author*[3]{\fnm{Hirotaka} \sur{Sato}}
\email{hirosato@ntu.edu.sg}

\author*[1,2]{\fnm{Masaki} \sur{Ogura}}
\email{oguram@hiroshima-u.ac.jp}

\affil[1]{\orgdiv{Graduate School of Information Science and Technology}, \orgname{Osaka University}, \orgaddress{\street{1-5 Yamadaoka}, \city{Suita}, \postcode{5650871}, \state{Osaka}, \country{Japan}}}

\affil[2]{\orgdiv{Graduate School of Advanced Science and Engineering}, \orgname{Hiroshima University}, \orgaddress{\street{1-3-2 Kagamiyama}, \city{Higashi-Hiroshima}, \postcode{7398511}, \state{Hiroshima}, \country{Japan}}}

\affil[3]{\orgdiv{School of Mechanical \& Aerospace Engineering}, \orgname{Nanyang Technological University}, \orgaddress{\street{50 Nanyang Avenue}, \city{Singapore}, \postcode{639798}, \country{Singapore}}}

\abstract{Cyborg insects refer to hybrid robots that integrate living insects with miniature electronic controllers to enable robotic-like programmable control. These creatures exhibit advantages over conventional robots in adaption to complex terrain and sustained energy efficiency. Nevertheless, there is a lack of literature on the control of multi-cyborg systems. This research gap is due to the difficulty in coordinating the movements of a cyborg system under the presence of insects' inherent individual variability in their reactions to control input. Regarding this issue, we propose a swarm navigation algorithm and verify it under experiments. This research advances swarm robotics by integrating biological organisms with control theory to develop intelligent autonomous systems for real-world applications.}

\maketitle

\section*{Introduction}\label{sec1}
The research on swarm navigation, referring to guiding a collective of agents to traverse an environment together, has captured increasing attention in recent years (\textit{1--6}). By harnessing the collective intelligence of autonomous entities, swarm navigation not only facilitates efficient traversal of unexplored terrains (\textit{7--9}) but also extends its utility to various sectors such as logistics (\textit{10, 11}), disaster response (\textit{12}), and agriculture (\textit{13}). In logistics, swarm navigation allows us to optimize route planning among multiple vehicles, enhancing transportation efficiency and reducing costs (\textit{10, 11}). In disaster response, it enables coordinated efforts among robotic teams to support disaster monitoring and survivor searching (\textit{12}). In agriculture, it supports precision farming techniques, monitoring crop health, and automating tasks to increase productivity and minimize environmental impact (\textit{13}).

However, using conventional robotic platforms such as UAVs and UGVs in swarm navigation has notable drawbacks, including their relatively large size and limited mobility. These drawbacks pose practical challenges, especially in constrained or crowded environments where spatial limitations impede effective task execution. In addition, their ability to operate in large terrains is directly constrained by the battery capacity, which limits endurance and overall efficiency.

This study introduces cyborg insects as a solution to the limitations of traditional robots in swarm navigation. Cyborg insects (Fig.~1a), which combine living insects with miniature electronic controllers, offer several advantages over conventional robots (\textit{14--18}). One of the advantages is their energy efficiency. Unlike robots, which typically rely on power-consuming drive mechanisms such as motors for locomotion, cyborgs exploit the natural mobility of insects and require less energy consumption (\textit{16, 18}). Cyborgs can also adapt to complex terrains (\textit{14, 15}). They can effortlessly maneuver around obstacles of various shapes and sizes and easily traverse through narrow spaces. Additionally, cyborgs are equipped with sophisticated sensing systems, which enable them to rapidly perceive and respond to the environment (\textit{17}).

Because of these features, research on cyborgs has recently gained increasing attention. Studies on cyborgs in (\textit{19, 20}) have mainly demonstrated their functionality in obstacle-free environments. Works in (\textit{21--23}) allow a single cyborg to navigate in rather complicated terrain by manually manipulating it along a preplanned path. In addition, the cyborg developed in (\textit{18}) can navigate to predetermined destinations and autonomously traverse unknown terrain. However, despite the aforementioned achievements in single-cyborg control, to the best of our knowledge, the multi-cyborg navigation problem has not been addressed in the literature (an illustration of a cyborg swarm can be seen in Figs.~1b, ~1c, and~1d).

A multi-agent system presents numerous advantages compared to a single-agent system, such as fault tolerance (\textit{24, 25}), distributed problem-solving (\textit{26, 27}), and applicability to large-scale problems (\textit{28}). Furthermore, the multi-cyborg system offers unique benefits, particularly in terms of enhanced robustness and intelligence. Regarding robustness, the interactions among the cyborgs in a multi-cyborg system can improve their ability to recover from an overturn because a well-designed swarm control algorithm can allow overturned cyborgs to be surrounded by neighbors that provide support points and aid in their rapid recovery. Additionally, the swarm behavior of a multi-cyborg system contributes to its intelligence.
When a cyborg encounters a potentially difficult area, it slows down and attempts to navigate through or around the obstacle. Neighboring cyborgs will maneuver around the slowed cyborg in the same way they would avoid obstacles. Consequently, this maneuvering leads the neighboring cyborgs to bypass the challenging area that caused the slowdown.
This cooperative behavior is similar to common human strategies, such as selecting shorter checkout lines in supermarkets for faster service. Furthermore, with appropriately designed swarm control algorithms, cyborgs that bypass the challenging zone can collectively assist the individuals trapped therein, resulting in the passage of the entire swarm.

While multi-cyborg systems offer various advantages over single-cyborg systems, controlling multiple cyborg insects presents many challenges. As partly biological entities, cyborg insects retain natural instincts that limit control precision compared to robots, with each insect responding differently to the same stimulus. Managing a single cyborg may be feasible despite this variability, but controlling a swarm presents a challenge. Advanced techniques such as adaptive control may be able to drive cyborgs to desired positions, but they often require intensive stimulation.
This level of stimulation increases the risk of habituation, where the responsiveness of cyborgs to control signals diminishes over time. The process is believed to involve synaptic changes in the nervous system, where the neurons become less responsive to the stimulus due to decreased neurotransmitter release or receptor sensitivity (\textit{29}).
Furthermore, certain innate behaviors of cyborg insects can even hinder the application of conventional swarm control logic. For instance, efforts to use stimulation for separating cyborg insects in close proximity often prove unsuccessful, frequently causing one insect to climb onto another, which results in entanglement. The entanglement presents a potential risk, as it may damage the cyborg’s backpack or its connection to the insect, making the cyborg non-functional. Collectively, these issues demonstrate that conventional control methods are not well-suited for managing cyborg swarms.

To address these challenges, we propose a swarm control algorithm specifically tailored for cyborg insects. This algorithm not only adapts to the unique characteristics of cyborg insects but also leverages these traits for effective swarm navigation. It offers improvements over conventional methods in terms of safety, endurance, and adaptability. Specifically, our algorithm harnesses the insects’ natural instincts, enabling efficient navigation through complex terrains without the need for precise locomotion control. The intermittent control inputs generated by our algorithm help mitigate habituation, thereby extending operational time. Additionally, our control strategy reduces the risk of entanglements among cyborgs, enhancing overall swarm safety. Experimental validation has confirmed the effectiveness of the proposed algorithm. Unlike previous studies that tested control algorithms in idealized settings, such as flat plains, our research demonstrates that both the control algorithm and the design of cyborg insects are adaptive and robust in navigating complex obstructed environments.

\section*{Results}
\subsection*{Swarm navigation algorithm design}\label{sec2}
The proposed algorithm in this paper can navigate a swarm of cyborgs from the start to a predetermined goal in an unknown sandy terrain in the presence of obstacles and hills (Fig.~2a). The multi-cyborg system to be controlled is composed of a leader and several followers, with agents capable of basic directional steering and forward motion. Each agent can detect neighbors within a limited sensing range and distinguish the leader from the followers. Only the leader is given the position of the goal. The proposed algorithm consists of two main components: motion planning and trajectory tracking (Fig.~2b). The motion planning algorithm provides the desired positions of cyborgs for the next time step based on their local information. The trajectory tracking algorithm then receives this information and computes the corresponding amplitude and types of stimulation (left, right, or acceleration) to be applied to the insects. 

Our motion planning algorithm was developed based on the observed behavior of tourists who follow a tour leader. Specifically, the algorithm consists of the following two clear and simple rules. The first rule is the free motion (FM) rule, which grants followers the freedom to move either when the leader is visible or when they remain close to the crowd. Otherwise, they adhere to the move-toward-crowd (MTC) rule, which guides followers toward the leader when visible and toward the direction of the crowd when not visible. Combining these rules, we propose the Tour Group Inspired (TGI) control algorithm. It can be applied to the multi-cyborg system: if the number of neighbors $m_{i}$ within follower~$i$'s free range (i.e., a circular area around it whose radius is smaller than its sensing range) is less than a threshold~$M$, follower~$i$ follows the MTC rule; otherwise, it makes free motions.
When the followers calculate $m_{i}$, the leader is given a higher weight than an individual follower. As a result, when the leader moves out of follower~$i$'s free range, the follower perceives the loss as equivalent to several neighbors, often falling below the threshold $M$. This triggers follower~$i$ to follow the leader, maintaining group cohesion and preventing the condition that the leader walks away while the group stays behind.

Note that an individual in free motion mode may not actively follow the swarm at all times. However, the control algorithm ensures that if the number of neighbors of a “free” follower falls under the threshold, it will be re-stimulated to rejoin the group. This mechanism allows the individual to temporarily engage in free motion while still maintaining overall cohesion with the swarm.
Moreover, this mechanism also mitigates the risk of entanglements that occur when cyborgs are stimulated in close proximity. If a cyborg insect is in an area with a sufficient number of neighboring insects, it is allowed to move freely without stimulation. In this unstimulated state, cyborgs rely on their natural instincts to maneuver around nearby insects, thereby avoiding entanglements.

We then present the details of our trajectory tracking algorithm design. We note that two types of stimulation, steering (left or right) and acceleration, can be applied to the insects. To determine the type of stimulation, the algorithm operates as follows. For each follower~$i$, the algorithm first divides its free range into several sectors. Then, the algorithm selects an arbitrary cyborg from the target sector (the one with the most neighbors) as the target cyborg, and steers the follower toward it. If angle~$\theta_{i}$ between the current moving direction of follower~$i$ and a line connecting itself with the target cyborg is less than a threshold~$\theta_{\textrm{threshold}}$, steering is not applied, and the algorithm proceeds to the acceleration decision. The rule that governs the acceleration is designed as follows. No acceleration will be applied if follower~$i$'s speed~$v_{i}$ exceeds a certain threshold~$v_{\textrm{threshold}}$. Otherwise, acceleration will be applied, and its magnitude is proportional to the distance between the follower and the target cyborg. This magnitude is bounded by the maximum voltage (2.5V) a cyborg can bear. Details of the control algorithm design are given in the supplementary materials.

Our TGI control algorithm capitalizes on the free motions of insects, potentially enhancing efficiency in obstacle and terrain negotiation. Additionally, the utilization of free motions decreases the frequency of electric stimulations on insects. Consequently, it reduces the likelihood of habituation of insects, saves the battery power of the backpack, and, in turn, prolongs the utilization of the cyborgs. Moreover, the FM rule in the TGI algorithm mitigates the impact of insects' individual variability on control performance. Specifically, the FM rule enables free motions of cyborgs in densely packed parts of the swarm, effectively averting entanglements and thereby bolstering the robustness of the multi-cyborg system.

\subsection*{Experimental verification of the proposed algorithm}\label{sec12}
The feasibility of the proposed swarm navigation algorithm is tested under real-world experiments in a proof-of-concept but challenging scenario. Ten trials of experiments were conducted to demonstrate the reproducibility of experiments.

The experiments used 20 cyborgs as the robot platform in a 3.5m by 3.5m sandy field with rocks and hills (Supplementary Fig. 1c). Among the 20 cyborgs, there is one leader, and the other 19 are the followers.
The leader agent was assigned a designated goal position, whereas the 19 follower agents were only provided the relative positions of their intermediate neighboring agents.
As aforementioned, the followers can distinguish between the leader and the followers. All the cyborgs have no information about obstacles and hills in advance. The leader was guided to the target using the algorithm proposed in our previous study  (\textit{18}), while the followers were controlled using the TGI algorithm developed in this work. Details of the experimental setup are given in the supplementary materials.

The proposed control strategy has been successfully implemented on the multi-cyborg system, effectively guiding the swarm to the designated goal area. To demonstrate the reproducibility of experiments, ten trials of experiments were conducted. We summarize the  corresponding results in Fig.~3. An illustration of the cyborg swarm navigation experiments in unknown obstructed terrain is provided in Fig.~3a. The trajectories of the cyborgs are presented in Fig.~3b, where the leader's path, the followers' trajectories, the positions where cyborgs were stimulated, and the final positions of all 20 cyborgs are presented with a yellow curve, blue curves, red circles, and black dots, respectively. Fig.~3c shows the corresponding top view of Exp~1. The results of Exp 2 to 10 are illustrated in Fig.~3d.
In Fig.~3d, a noticeable deviation of the leader in Exp 2 is observed, where it appears to jump to a different position and then return to its designated trajectory. This is due to the leader's marker, which is utilized for localization within the VICON motion capture system, becoming contaminated and temporarily untrackable for more than 5 seconds. During this period, the control algorithm was unable to monitor the leader's precise position. The following mechanism was designed to mitigate such disruptions. When the leader’s marker becomes untrackable, the followers navigate toward the leader's last known position, ensuring group cohesion and preventing the group from dispersing. Once the VICON system reacquires the leader's marker, the control algorithm re-engages, guiding the leader back to its intended trajectory, at which point the follower agents resume real-time tracking.

The degree of autonomy depicted in Fig.~3e illustrates the efficacy of the proposed algorithm in leveraging insects' instincts. In this study, we define the degree of autonomy for the multi-cyborg system as follows: For each experimental trial, we calculate the ratio~$A^{l}_{k}$ for the leader and average ratio $A^{f}_{k}$ for the followers in trial $k$, between the time without stimulation and the whole experimental period. Then, the degree of autonomy of the total 10 trials of experiments is defined as $\Sigma_{k=1}^{10} A^{l}_{k} /10$ for the leader, and~$\Sigma_{k=1}^{10} A^{f}_{k}/10$ for the followers. As illustrated in Fig.~3e, the blue bar indicates the degree of autonomy of the followers, while the yellow bar represents the degree of autonomy of the leaders. The error bars denote the standard deviation.
Under conventional control, the leader's degree of autonomy averaged 0.26 across 10 experimental trials, while the followers, under the proposed control algorithm, attained an average degree of autonomy of 0.5. Our approach nearly doubles the cyborgs' free-motion time, effectively mitigating the likelihood of habituation of insects, saving the battery power of the backpack, and in turn, prolonging the utilization of the swarm system.

\section*{Discussion}\label{sec11}
During the experiments, several other intriguing phenomena demonstrated the features of the proposed TGI control algorithm:
\begin{itemize}
    \item The employment of free motions reduces entanglements among cyborgs, thus leading to a higher safety of the multi-cyborg system.
    \item The interactions among neighbors may facilitate a trapped cyborg to escape a difficult situation.
    \item Neighbors can help a cyborg recover from an overturn, which, in turn, enhances the robustness of a cyborg swarm.
\end{itemize}
These features are detailed as follows.

First, the TGI control algorithm improves the safety of the multi-cyborg system by reducing entanglement between cyborgs. Our experiments showed that cyborgs instinctively avoid collisions with neighbors during free movement. However, when two cyborgs approached each other too closely and the separation rule from the conventional BOIDS algorithm (\textit{30}) was applied to stimulate the cyborgs to move away from each other, they failed to separate and instead became entangled. Entanglement can lead to undesirable results, such as damage to the cyborgs. This problem was often encountered with the conventional control based on the BOIDS algorithm when neighbors became too close.
As depicted in Fig.~4a, the central line within the box indicates the median entanglement number, while the boundaries represent the 25th and 75th percentiles (lower and upper quartiles). The whiskers extend to the smallest and largest entanglement numbers within 1.5 times the interquartile range from the quartiles. This figure demonstrates that the proposed TGI algorithm reduces entanglements compared to the BOIDS algorithm. Fig.~4b depicts entanglements where multiple cyborgs overlap, whereas Fig.~4c shows closely positioned but non-entangling cyborgs. The feasibility of using free motion to prevent entanglement is demonstrated in Fig.~4d and Fig.~4e. When the cyborgs approached each other, the BOIDS method tried to separate them by applying intensive stimulation, which proved to be unsuccessful (Fig.~4d). Conversely, the proposed TGI control exploited the free motion of the cyborgs, taking advantage of their innate behaviors to avoid entanglement (Fig.~4e).

Second, the interactions among neighbors can help a trapped cyborg escape a difficult situation. As shown in Fig.~5a, in an experiment, a cyborg's ``Y'' shape marker became wedged on the edge of an obstacle. Meanwhile, following the proposed control algorithm, other cyborgs were navigated around the cyborg, which indirectly let the swarm bypass the obstacle. Furthermore, it is noteworthy that, in line with the MTC rule of the proposed control algorithm, cyborgs bypassing challenging zones can assist trapped cyborgs, facilitating the passage of the entire swarm. As illustrated in the snapshots, when the cyborg marked by a yellow circle initially became wedged on an obstacle, it remained unstimulated due to its proximity to the neighbors. However, when the neighbors were bypassing the stuck one ($t=t_2$), they started to ``attract'' it (stimulations were reapplied to the cyborg), pulling it away from the obstacle.

Third, neighbors can aid a overturned cyborg for more efficient uprighting.
Although neighboring cyborgs may navigate around an overturned cyborg in a manner similar to how they avoid obstacles, they also play a passive yet crucial role in facilitating its recovery. The overturned cyborg can utilize interactions with nearby teammates, such as physical contact or grasping, to aid in up-righting itself. This cooperative interaction enhances the overall resilience of the swarm without requiring direct, algorithmic intervention.
The bar chart in Fig.~5b illustrates three cases of self-attempted and neighbor-aided recovery from overturns. The discrepancy in recovery performance is particularly evident in Case 3, where a cyborg struggled for 26.5s before grasping a passing neighbor and then successfully recovered in only 4.5s. Snapshots of Cases 1 and 3, where cyborgs recovered with the help of neighbors, are also provided in Fig.~5b, respectively. The proposed TGI control algorithm ensures the presence of neighbors around each cyborg and, in turn, enhances the robustness of the multi-cyborg system.

In summary,
our approach offers advantages over conventional control methods in terms of safety, endurance, and adaptability when applied to cyborg swarms. Without relying on precise locomotion control, our algorithm achieves a compact swarm through adjustable parameter design and effectively harnesses insects’ instincts for efficient obstacle negotiation in complex, obstructed terrains. Additionally, our algorithm mitigates the risk of habituation and extends operational time by reducing the need for control inputs by an average of $50\%$ compared to the individual control. This reduction is accomplished through the free-motion and move-toward-crowd rules, which leverage the insects’ natural motion while maintaining swarm cohesion. Furthermore, our algorithm mitigates the risk of entanglements that can destabilize the swarm and damage the cyborgs, using a switching mechanism between stimulated and free motions based on local swarm compactness. Experiments demonstrate that this design reduces the number of entanglements by over $85\%$ on average compared to the BOIDS method, enhancing the overall safety of the swarm system.

Note that although the control algorithm proposed in this work is theoretically decentralized, the cyborgs in our experiments were provided data from a centralized motion capture system.
To achieve decentralized sensing, enabling cyborgs to directly measure distances between each other, we propose a hybrid approach for future work that combines Inertial Measurement Unit (IMU) and Ultra-Wideband (UWB) technologies. Specifically, each cyborg insect will be equipped with a low-power IMU to measure yaw, pitch, and roll angles. This data will be used to estimate position by analyzing body vibrations during locomotion, with speed derived from step frequency and position calculated through time integration. To mitigate IMU drift, UWB modules will be deployed to periodically update each insect’s position using precise distance measurements. This decentralized hybrid IMU-UWB system can provide robust, accurate localization for the cyborgs without a motion capture system.
It can be a promising direction for the practical utilization of the proposed control strategy.

\section*{Methods}

\subsection*{Control algorithm design}
Let us define the position of a cyborg by $\textbf{q}_{i}(t)\in\mathbb{R}^{2}$, velocity by $\textbf{V}_{i}(t)\in\mathbb{R}^{2}$, and orientation by a unit vector $\textbf{o}_{i}(t)\in\mathbb{R}^{2}$, at time $t$.
Without loss of generality, we assign $i=0$ for the leader and $i \neq 0$ for followers.
For cyborg $i$ at time $t$, we define the sets of its neighbors by
\begin{equation}
  \mathcal{N}_i(t) = \{j \in \{0, \ldots , N\}\backslash \{i\} \ |\  0 < \|\textbf{q}_{j}(t)-\textbf{q}_{i}(t)\| \leq r_{s} \},
\end{equation}
where $N$ is a positive integer and $r_{s}$ is a positive constant, denoting the radius of the sensing range. Similarly, we define
\begin{equation}
  \mathcal{M}_i(t) = \{j \in \{0, \ldots , N\}\backslash \{i\} \ |\  0 < \|\textbf{q}_{j}(t)-\textbf{q}_{i}(t)\| \leq r_{\textrm{free}} \},
\end{equation}
where positive constant $r_{\textrm{free}}$ denotes the radius of the free range. Thus
\begin{equation}
    m_{i}=|\mathcal{M}_i|,
\end{equation}
is the number of neighbors in the free range of cyborg $i$. Note that $r_{\textrm{free}} < r_{s}$.

Based on these definitions, the swarm control algorithm is detailed as follows. It consists of two rules:
\begin{itemize}
    \item FM rule: for cyborg $i \in \{1, \ldots, N\}$ (followers), if the number of neighbors within the free range $m_{i} > M$ where $M$ is a positive integer, it engages in free motions; otherwise, it follows the rule below.
    \item MTC rule: for cyborg $i \in \{1, \ldots , N\}$, its sensing range is divided into s sectors, and steered to the target sector (the one with the largest number of neighbors where the leader is counted as $N$ cyborgs). On the other hand, the leader $ i = 0 $ is driven to the goal position. The aforestated functions are realized by
    \begin{equation}
        \textbf{f}_{m, i}(t)=\begin{cases}
        \textbf{p}^{\textrm{goal}}(t)-\textbf{q}_{i}(t), & i = 0, \\
        \textbf{p}^{\textrm{target}}(t)-\textbf{q}_{i}(t), & i \in \{1, \ldots , N\},
        \end{cases}
    \end{equation}
    where $\textbf{p}^{\textrm{goal}}(t)$ and $\textbf{p}^{\textrm{target}}(t)$ denotes the position of a target cyborg (an arbitrary cyborg in the target sector) and the goal for the leader, respectively. Then, the desired orientation of cyborg $i$ can be computed by
    \begin{equation}
        \textbf{o}_{i}^{\textrm{des}}(t) = \mathbf{\phi}(\textbf{f}_{m, i}(t)),
    \end{equation}    
    where $\phi:\mathbb{R}^{2}\backslash \{\textbf{0}\} \rightarrow \mathbb{R}^{2}\backslash \{\textbf{0}\}: \textbf{x} \mapsto \frac{\textbf{x}}{\|\textbf{x}\|}.$

    Note that the following repulsion function can also be integrated into the MTC rule to achieve a rather sparse swarm: cyborg $i$ is steered away from a neighbor $j$ at time $t$, according to
    \begin{equation}
        \textbf{f}_{r, i}(t)=\begin{cases}
        \textbf{0}, & i = 0, \\
        - \sum_{j\in\mathcal{N}_{i}(t)} \frac{1}{\|\textbf{q}_{j}(t)-\textbf{q}_{i}(t)\|^2}(\textbf{q}_{j}(t)-\textbf{q}_{i}(t)), & i \in \{1, \ldots , N\}.
        \end{cases}
    \end{equation}
    Combining these functions, the desired velocity of follower $i$ is computed according to
\begin{equation}
  \textbf{V}_{i}^{\textrm{des}}(t)=k_{m}\textbf{f}_{m, i}(t) + k_{r}\textbf{f}_{r, i}(t),  
\end{equation}
where $k_{m}$, $k_{r}$ are constant weights. Then, the desired position $\textbf{q}_{i}^{\textrm{des}}(t)$ of cyborg $i$ can be computed by
\begin{equation}
    \textbf{q}_{i}^{\textrm{des}}(t) = \textbf{q}_{i}(t)+\textbf{V}_{i}^{\textrm{des}}(t)\Delta t,
\end{equation}
where $\Delta t$ denotes the time step.
    
\end{itemize}

Thus, one defines the position error of cyborg $i$ by
$
    \textbf{e}_{q,i}(t)=\textbf{q}_{i}^{\textrm{des}}(t)-\textbf{q}_{i}(t)
$
and the orientation error by
$
    \textbf{e}_{o,i}(t)=\textbf{o}_{i}^{\textrm{des}}(t)-\textbf{o}_{i}(t)
$.
The angle between $\textbf{o}_{i}^{\textrm{des}}(t)$ and $\textbf{o}_{i}(t)$ is denoted by $\theta_{i}$, and the speed of agent $i$ is represented by $v_{i}$.
The computed $\textbf{e}_{q,i}(t)$, $\textbf{e}_{o,i}(t)$ and the measured $\theta_{i}$, $v_{i}$ are passed to the trajectory tracking algorithm summarized in \textbf{Algorithm 1} in the Supplementary Information.

\section*{Data availability}
The data that support the findings of this study are available within the paper and Supplementary Information files. Source data are provided in this paper.

\section*{Code availability}
The code that supports the findings of this study is available within the Supplementary Information files.

\section*{Acknowledgements}
The authors received funding from JST (Moonshot R\&D Program, Grant Number JPMJMS223A).
\section*{Author contributions} N. W., H. S., and M. O. conceived and designed the research. Y. B., M. O., and N. W. developed the swarm control algorithm. P. T. T. N., H. D. N., D. L. L., and K. K. built the experimental platform. Y. B., P. T. T. N., Q. H. H., K. K., and Y. X. S. T. prepared the cyborg insects. Y. B., P. T. T. N., H. D. N., Q. H. H, Y. D., and Y. X. S. T. collected the raw data for the cyborg experiments. Y. B., J. S., and Y. D. conducted the data analysis. Y. B., J. S., P. T. T. N., M. O., N. W., and H. S. contributed to the preparation of the manuscript. N. W., H. S., and M. O. supervised the research.
\section*{Competing interests} The authors declare no competing interests.

\section*{Figures}

\begin{figure}[H]
  \centering
  \includegraphics[width=12cm]{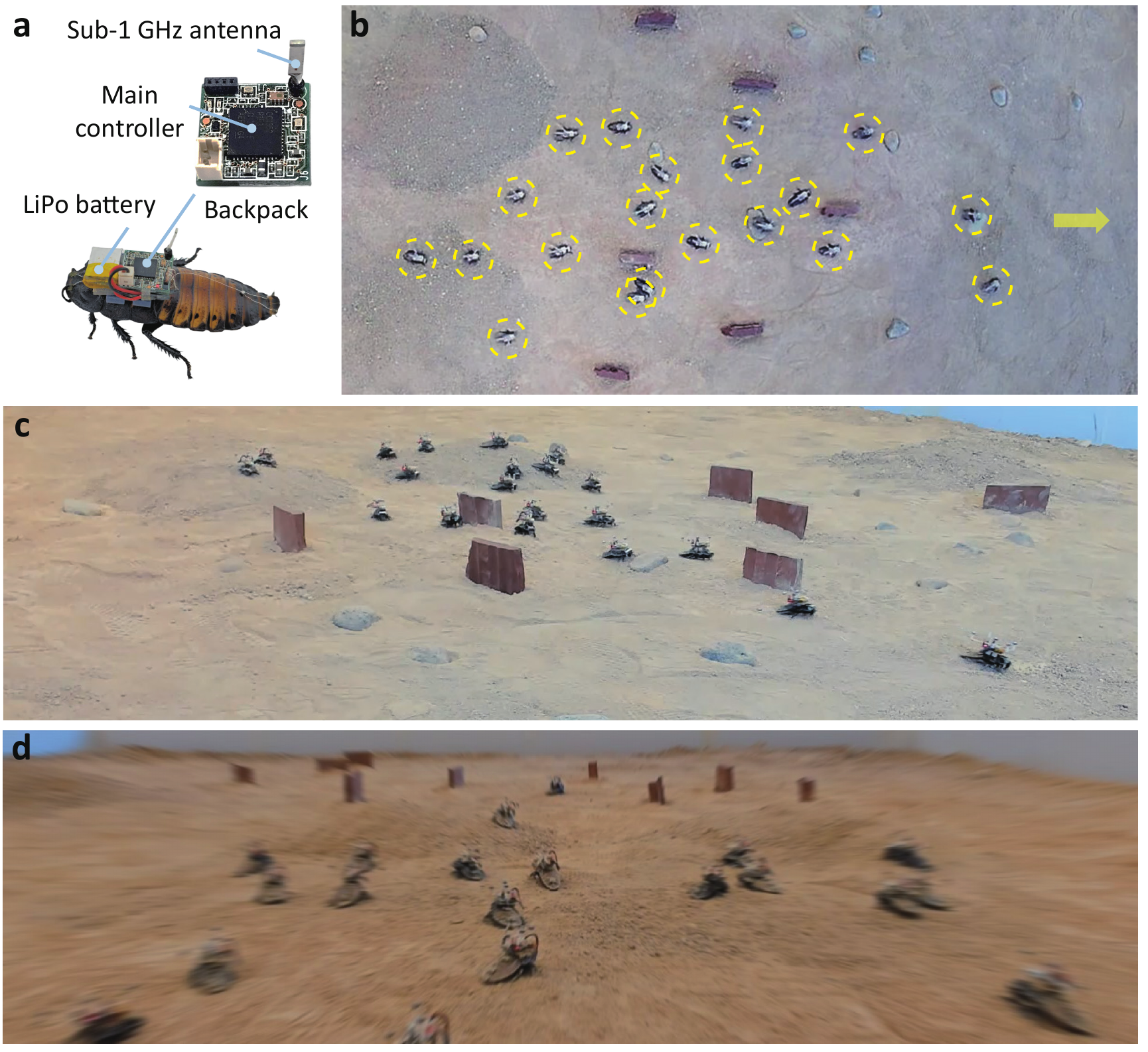}
  \caption{\textbf{Overview of the cyborg swarm navigation.} (\textbf{a}) A cyborg insect. In this work, the Madagascar hissing cockroach (\textit{Gromphadorhina portentosa}) was chosen for the cyborg system. A backpack circuit board attached to the insect was designed to house the necessary systems, including the locomotion control system and wireless communication module. It was powered by a rechargeable LiPo battery. How to prepare a cyborg insect is detailed in the supplementary materials. (\textbf{b}) An illustration of cyborg swarm navigation (top view). A decentralized algorithm was proposed to navigate the cyborg swarm to a designated goal through a sandy area in the presence of hills and obstacles. All the cyborgs have no information about obstacles and hills in the field. (\textbf{c}) The side view and (\textbf{d}) the back view of a cyborg swarm. A comprehensive presentation of this work can be found in Supplementary Video 1.}
  \label{intro}
\end{figure}

\begin{figure}[H]
  \centering
  \includegraphics[width=12.5cm]{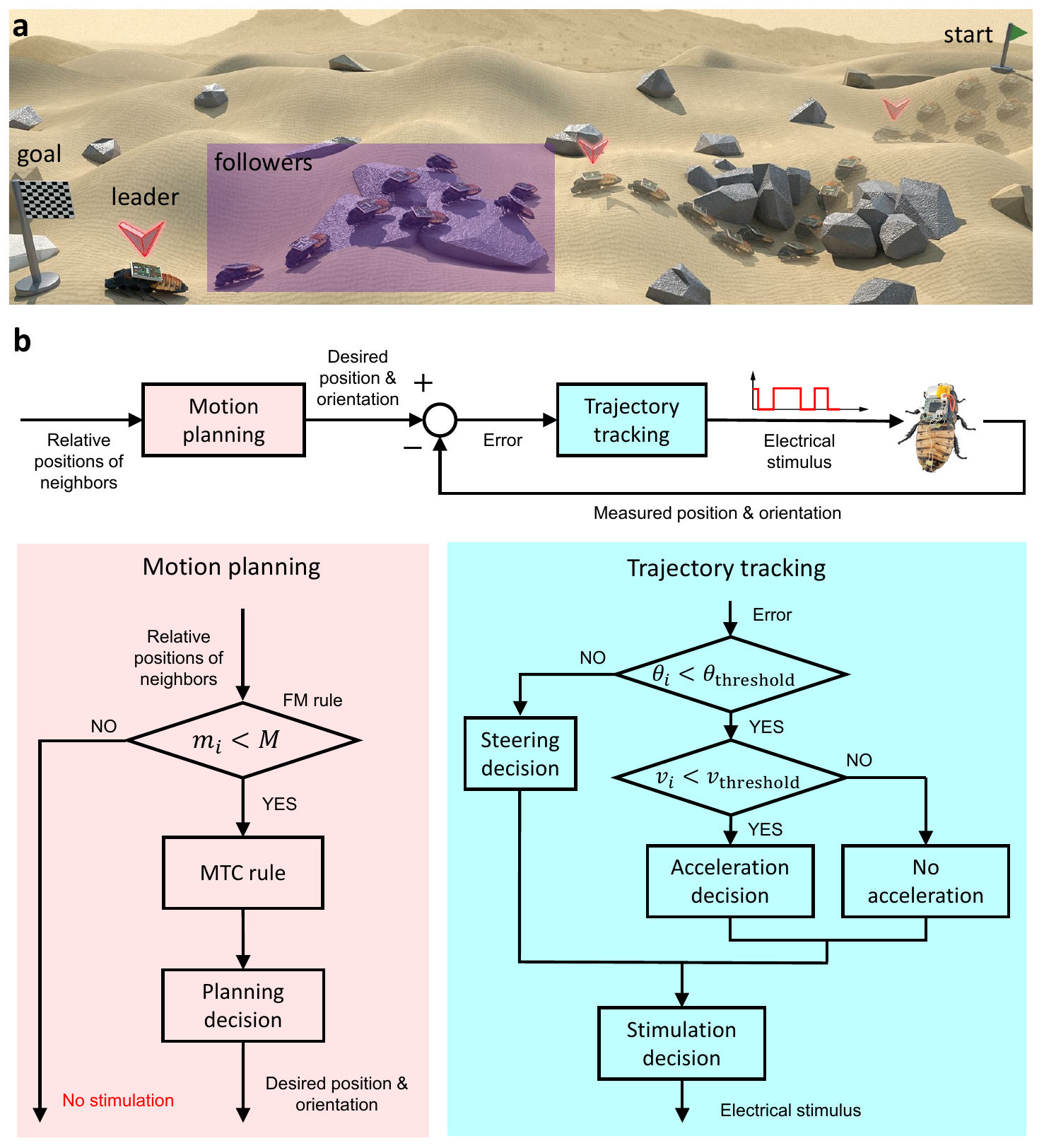}
  \caption{\textbf{Navigation algorithm design for a cyborg swarm.} (\textbf{a}) The cyborg swarm consists of a leader and several followers. Each cyborg can detect neighbors within a limited sensing range, and distinguish the leader and followers. Only the leader is given the position of the goal. (\textbf{b}) An illustration of the proposed navigation algorithm. It consists of two parts: motion planning and trajectory tracking. The motion planning algorithm provides the desired positions of cyborgs for the next step based on their local information and passes it to the trajectory-tracking algorithm. The trajectory tracking algorithm computes the corresponding amplitude and types of stimulation (left, right, or acceleration) to be applied to the insects.}
  \label{control}
\end{figure}

\begin{figure}[H]
  \centering
  \includegraphics[width=9.5 cm]{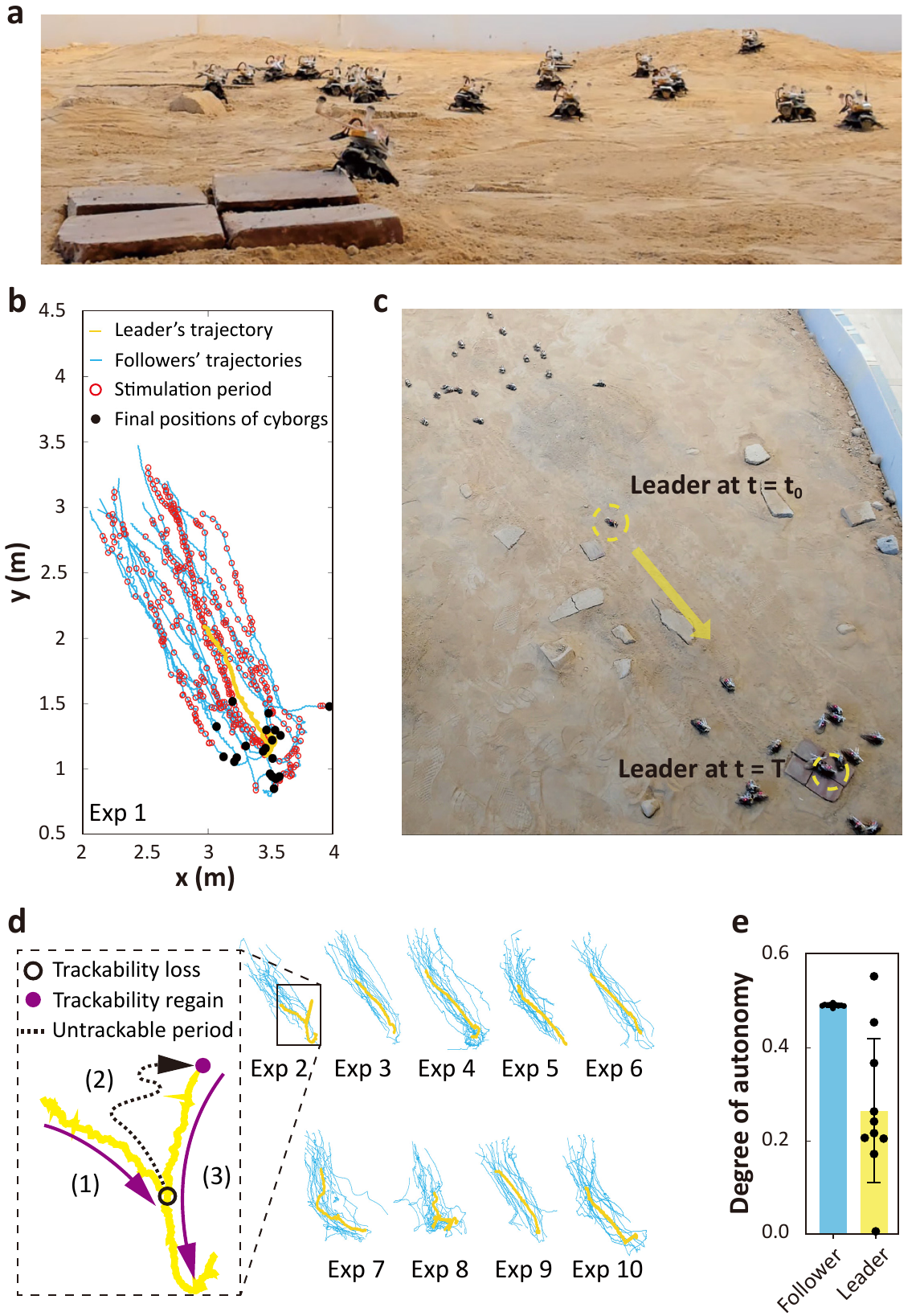}
  \caption{\textbf{Summary of swarm navigation experiments.} (\textbf{a}) An illustration of experiments in an obstructed soft terrain (front view). (\textbf{b}) Path plot of Exp~1.
  (\textbf{c}) The top view of Exp~1. The video of Exp~1 is Supplementary Video 2. (\textbf{d}) Path plots of Exp~2 to~10. In Exp~2, the leader’s marker was untrackable for more than 5s, but the followers maintained group cohesion by moving to the leader's last known position. They resumed tracking the leader after it gained its trackability. (\textbf{e}) The degree of autonomy for the leader under conventional control (yellow) and the followers under our proposed control algorithm (blue) in 10 trials of experiments. The error bars denote the standard deviation. Our approach nearly doubles the cyborgs' free-motion time, effectively mitigating the likelihood of habituation of insects. Source data are provided as a Source Data file.}
 \label{result}
\end{figure}

\begin{figure}[H]
  \centering
  \includegraphics[width=8 cm]{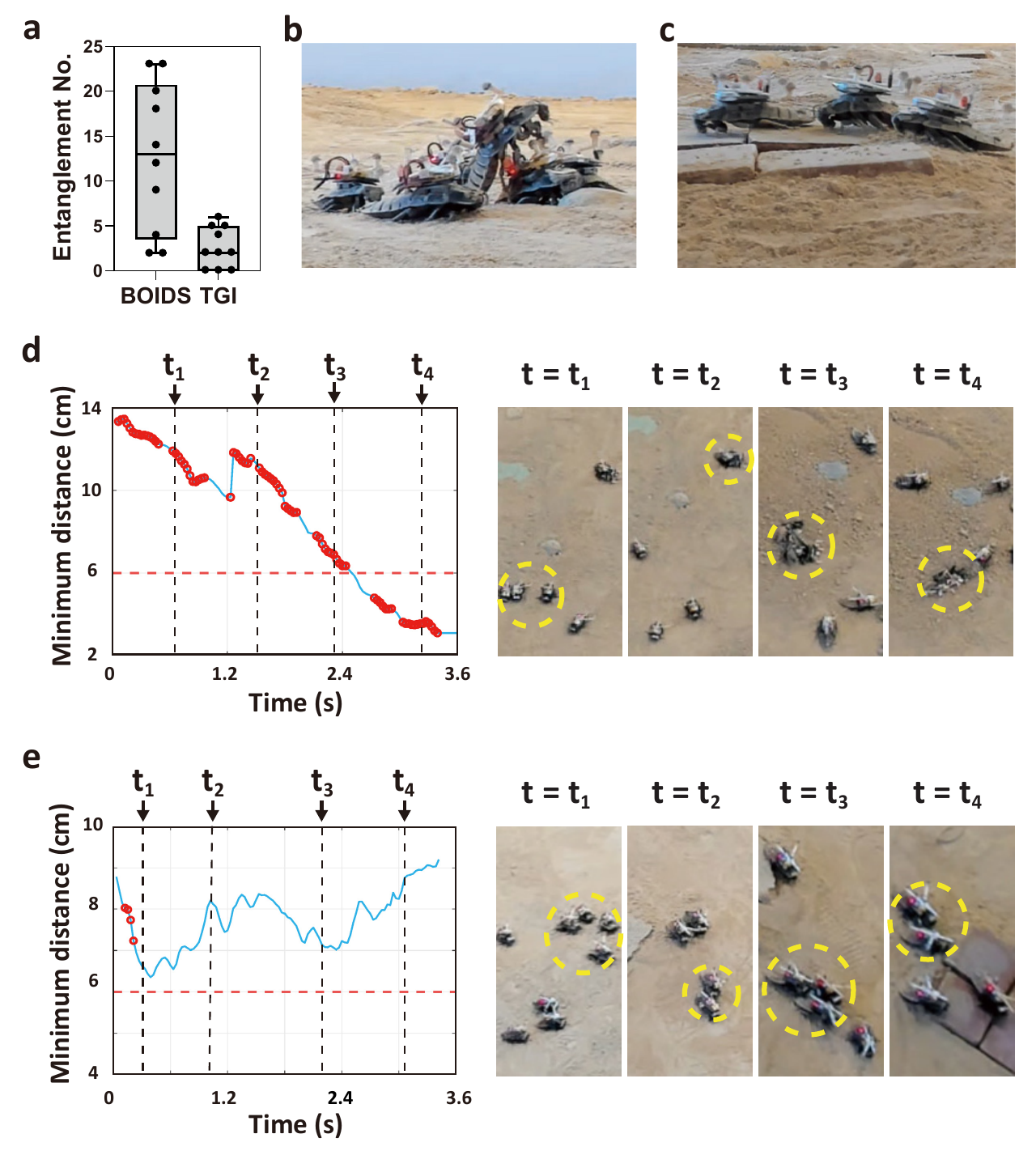}
  \caption{\textbf{Comparison between the experimental results under the conventional BOIDS and the proposed TGI algorithms.} (\textbf{a}) The frequency of entanglements in 10 trials of experiments, respectively, under two control algorithms. The central line within the box indicates the median entanglement number, while the boundaries represent the 25th and 75th percentiles (lower and upper quartiles). The whiskers extend to the smallest and largest entanglement numbers within 1.5 times the interquartile range from the quartiles. The number of entanglements under the proposed TGI algorithm is notably lower than that under the BOIDS algorithm. (\textbf{b}) An illustration of entanglements: when parts of two or more cyborgs overlap with each other. Entanglement can lead to undesirable results, such as damage to the cyborgs. (\textbf{c}) A group of close but non-entangling cyborgs. (\textbf{d}) The minimum distance among all pairs of cyborgs circled out in experimental snapshots in one trial of experiments conducted with the BOIDS algorithm. Red points indicate instances where at least one of those cyborgs received electrical stimulation. The frequent stimulations on cyborgs that were too close to each other led to entanglements (distance less than 6 cm). (\textbf{e}) The minimum distance in one trial of experiments under TGI control. The TGI algorithm leverages insects' instincts to prevent entanglements. The corresponding snapshots reveal that cyborgs can move closely without entanglements under TGI control. The experiment results are shown by Supplementary Video 3. Source data are provided as a Source Data file.}
  \label{boids_tgi}
\end{figure}

\begin{figure}[H]
  \centering
  \includegraphics[width=12cm]{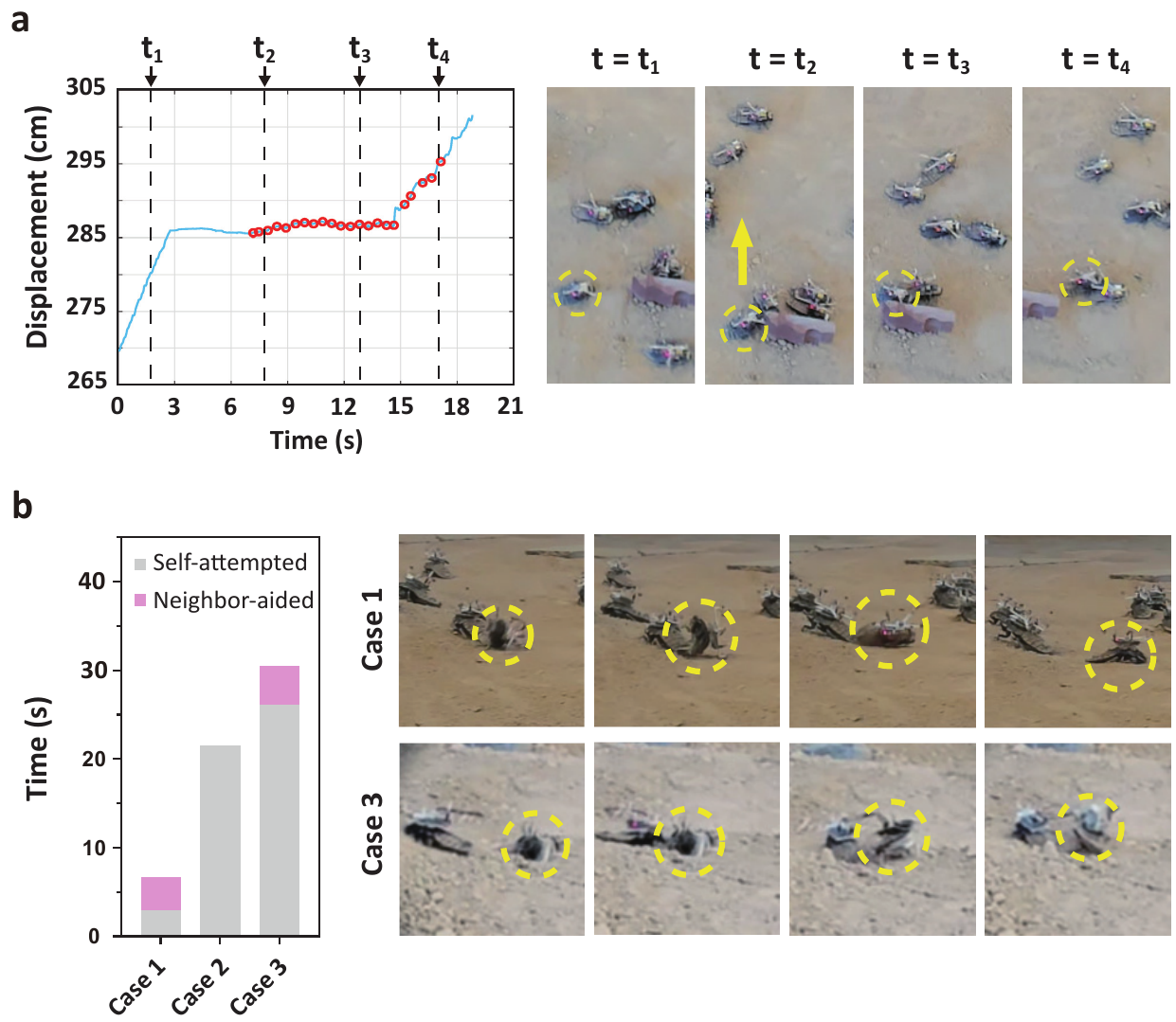}
  \caption{\textbf{The process of an immobilized cyborg being saved by neighbors.} (\textbf{a}) Entangle and detach from obstacles. A cyborg's ``Y'' shape marker became wedged on an obstacle ($t = t_2$), and the corresponding displacement almost remained the same during this period. Then the surrounding cyborgs navigated around the trapped one and the obstacle ($t = t_3$). Through our algorithm, the trapped cyborg gradually overcame the obstacle with the ``attractive force'' from other cyborgs, and the displacement started changing again. The red dots in the displacement plot represent electrical stimulations applied to the cyborgs. Finally, the cyborg detached from the obstacle ($t = t_4$). See Supplementary video 4. (\textbf{b}) Recover from an overturn. The grey bar in the bar chart denotes the time interval of a self-attempted period for recovery of an overturned cyborg. The bar's bottom and top indicate the start and ending moments. Similarly, the purple bar denotes the time taken for recovery with help from a neighbor. In Cases 1 and 3, the overturned cyborg first attempted and failed to recover by itself for a while, then successfully recovered by grabbing a passing neighbor. The corresponding recovery processes are illustrated by the snapshots. The experiment results are shown by Supplementary Video 5. Source data are provided as a Source Data file.}
  \label{kazhu}
\end{figure}

\end{document}